\documentclass[runningheads]{llncs}
\usepackage{graphicx}
\usepackage{dsfont}
\usepackage{amsmath}
\usepackage{amsfonts}
\usepackage{caption}
\usepackage{subcaption}
\usepackage{booktabs}
\usepackage[title]{appendix}
\usepackage{float}
\usepackage[sorting=none]{biblatex}
\usepackage[hidelinks]{hyperref}
\bibliography{lite}
\begin{document}
\title{Random Noise vs State-of-the-Art Probabilistic Forecasting Methods
 : A case study on CRPS-Sum discrimination ability}
\titlerunning{A case study on CRPS-Sum discrimination ability}
%
\author{Alireza Koochali\inst{1,2,3}\orcidID{0000-0001-7370-9369} \and
Peter Schichtel\inst{1}\orcidID{0000-0002-7634-4964} \and
Andreas Dengel\inst{2,3}\orcidID{0000-0002-6100-8255}\and
Sheraz Ahmed\inst{2}\orcidID{0000-0002-4239-6520}}
\authorrunning{A. Koochali et al.}
%
\institute{IAV GmbH, Carnotstr. 1, 10587 Berlin, Germany\\ 
\email{\{alireza.koochali,peter.schichtel\}@iav.de} \and
 DFKI GmbH, Trippstadter Str. 122, 67663 Kaiserslautern, Germany\\
\email{\{alireza.koochali,andreas.dengel,sheraz.ahmed}@dfki.de\} \and
University of Kaiserslautern, Erwin-Schrödinger-str. 52, 67663 Kaiserslautern, Germany\\
\email{akoochal@rhrk.uni-kl.de}}
\maketitle              
\begin{abstract}
The recent developments in the machine learning domain have enabled the development of complex multivariate probabilistic forecasting models. Therefore, it is pivotal to have a precise evaluation method to gauge the performance and predictability power of these complex methods.  To do so, several evaluation metrics have been proposed in the past (such as Energy Score, Dawid-Sebastiani score, variogram score), however, they cannot reliably measure the performance of a probabilistic forecaster. Recently, CRPS-sum has gained a lot of prominence as a reliable metric for multivariate probabilistic forecasting. This paper presents a systematic evaluation of CRPS-sum to understand its discrimination ability. We show that the statistical properties of target data affect the discrimination ability of CRPS-Sum. Furthermore, we highlight that CRPS-Sum calculation overlooks the performance of the model on each dimension. These flaws can lead us to an incorrect assessment of model performance. Finally, with experiments on the real-world dataset, we demonstrate that the shortcomings of  CRPS-Sum provide a misleading indication of the probabilistic forecasting performance method. We show that it is easily possible to have a better CRPS-Sum for a dummy model, which looks like random noise, in comparison to the state-of-the-art method.

\keywords{Time-series analysis, Probabilistic forecasting, Assessment}

\end{abstract}

\section{Introduction}

In the last decades, probabilistic forecasting has drawn a lot of attention in the scientific community which led to the fast-paced development of new methods as well as application in a wide variety of domains including renewable energies~\cite{pinson2013wind,bacher2009online,chen2018model}, weather forecast~\cite{cloke2009ensemble,racah2016extremeweather,rodrigues2018deepdownscale}, seismic hazard prediction~\cite{mousavi2019cred,ross2019phaselink}, and health care~\cite{avati2018improving}. Due to the recent upsurge in Deep Learning, several approaches have been proposed which are based or inspired by these powerful neural network. To name some of these techniques, there are DeepAR~\cite{salinas2020deepar}, Low-Rank Gaussian Copula Processes~\cite{salinas2019high}, Conditioned Normalizing Flows~\cite{rasul2020multi}, Normalizing Kalman Filter\cite{de2020normalizing}, Denoising Diffusion Model~\cite{rasul2021autoregressive}, and Conditional Generative Adversarial Networks~\cite{koochali2019probabilistic,koochali2020if}. In many proposed solutions, we do not have direct access to the probability distribution of the model over future values. Hence, we acquire the forecast through Monte Carlo sampling. Once these models are available, it is important to have evaluation methods which can be used to gauge the performance. Assessment of probabilistic models poses a special challenge when we only have access to forecasts samples from the probabilistic model.\medskip

\noindent
Garthwaite et al.~\cite{garthwaite2005statistical} coined the concept of scoring rules for summarizing the quality of probabilistic forecaster with a numerical score~\cite{gneiting2007strictly}. A scoring rule is expected to make careful assessment and be honest~\cite{garthwaite2005statistical}. Gneiting et al.~\cite{gneiting2007strictly} proposed Continuous Ranked Probability Score(CRPS) for univariate and Energy Score (ES) for multivariate time series as strictly proper scoring rules. While CRPS presents a robust and reliable assessment method for univariate time series forecasting, ES discrimination ability diminishes in higher dimensionalities~\cite{pinson2013discrimination}. Several other multivariate metrics~\cite{pinson2013discrimination,scheuerer2015variogram} were proposed to address probabilistic forecaster assessment in higher dimensions however, each of them had a flaw that makes them unsuitable for the assessment task. For instance, variogram score~\cite{scheuerer2015variogram} is a proper scoring rule which can reflect the misalignment in correlation very well, but it lacks strictness property. Dawid-Sebastiani score~\cite{dawid1999coherent} employs only the first two moments of distribution for evaluation which is not sufficient for many applications. A thorough analysis of these metrics is provided in~\cite{ziel2019multivariate}.\medskip

\noindent
Recently, Salinas et al.~\cite{salinas2019high} suggest CRPS-Sum as a new proper multivariate scoring rule. This scoring rule has been well-received in the scientific community~\cite{salinas2019high,rasul2020multi,rasul2021autoregressive,de2020normalizing}. The properties of CRPS-Sum are not studied so far. \medskip

\noindent
In this paper, our goal is to discuss the discrimination ability of CRPS-Sum. We have conducted several experiments on artificial and real dataset to investigate quantification power of CRPS-sum for the performance of a probabilistic forecaster. Based on the experiments' results, we point out the loopholes in this metric and discuss how CRPS-Sum can mislead us in interpretation of a model's performance.\medskip

\section{Problem Specification}
\label{sec:2}
The forecasting task deals with predicting the future given historical information of a time series. A time series can have multiple dimensions. The notation $x^{i}_{t}$ indicates the value of a time series at the time-step $t$ in the $i-th$ dimension. If a time series has only one dimension, it is called a univariate time series, otherwise, it is a multivariate time series.\medskip

\noindent
In the forecasting task, given $x_{0:T}$ as historical information, we are interested in predicting values for $x_{T+1:T+h}$ where h stands for the horizon of forecast. In probabilistic forecasting, the target is to acquire the range of possible outcomes with their corresponding probabilities. In more concrete terms, we aim to model the following conditional probability distribution:
\begin{equation}
    P(x_{T+1:T+h}|x_{0:T}).
    \label{eq:prob_forcast}
\end{equation}
For the assessment of a probabilistic forecasting model, the goal is to measure how well a model is aligned with the probability distribution of the data. In other words, we want to calculate the divergence between $P_\text{model}$ and $P_\text{data}$.

\section{Evaluation Metrics for Probabilistic Forecasting Models}

Our first challenge for assessing a probabilistic model is that, in real-world scenarios, we do not have access to the true generative process distribution i.e.  $P_\text{data}$. We only have access to the observations from $P_\text{data}$. The scoring rule provides a general framework for evaluating the alignment of $P_\text{data}$ with $P_\text{model}$. A scoring rule is any real-valued function which provides a numerical score based on the predictive distribution (i.e. $P_\text{model}$) and a set of observations $X$.
\begin{equation}
    S(P_\text{model},X)
\end{equation}
The scoring can be defined positively or negatively orientated. In this paper, we consider the negative orientation, since it can be interpreted as the model error and as a result, it is more popular in the scientific community. Hence, the lower score indicates a better probabilistic model.\medskip

\noindent
A scoring rule is proper if the following inequality holds:
\begin{equation}
    S(P_\text{model},X) \geq S(P_\text{data},X)
    \label{eq:proper}
\end{equation}
A scoring rule is strictly proper if the equality in equation~\ref{eq:proper} hold if and only if $P_\text{model}=P_\text{data}$~\cite{gneiting2007strictly}. Therefore, only the model which is perfectly aligned with the data generative process can acquire the lowest strictly proper score. Various realization of scoring rules have been proposed to evaluate the performance of a probabilistic forecaster. Below, we review three scoring rules which are commonly used for the assessment of a probabilistic foresting model.

\subsection{Continuous Ranked Probability Score (CRPS) }
CRPS is a univariate strictly proper scoring rule which measures the compatibility of a cumulative distribution function $F$ with an observation $x \in \mathbb{R}$ as
\begin{equation}
    \operatorname{CRPS}(F, x)=\int_{\mathbb{R}}(F(y)-\mathds{1}\{x \leq y\})^{2} \mathrm{~d} y \,,
    \label{eq:crps_F}
\end{equation}
where $\mathds{1}\{x \leq y\}$ is the indicator function, which is one if $x \leq y$ and zero otherwise.\medskip

\noindent
The predictive distributions are often expressed in terms of samples, possibly through Monte Carlo sampling~\cite{gneiting2007strictly}. Fortunately, there are several methods to estimate CRPS given only samples from predictive distribution. The precision of these approximation methods depends on the number of samples we use for estimation. Below you can find a list of the most used techniques.\medskip\\

\paragraph{Empirical CDF:}
 In this technique, we approximate the CDF of a predictive model using its samples.
\begin{equation}
\hat{F}(y)=\frac{1}{n} \sum_{i=1}^{n} \mathds{1}\left\{x_{i} \leq y\right\}\,.
\label{eq:crps_cdf}
\end{equation}
Then, we can use $\hat{F}(y)$ in conjunction with Equation~\ref{eq:crps_F} to approximate CRPS.
\paragraph{Quantile-based:} The pinball loss or quantile loss at a quantile level $\alpha \in [0, 1]$ and with a predicted $\alpha^\text{th}$ quantile $q$ is defined as 
\begin{equation}
    \Lambda_{\alpha}(q, x)=\left(\alpha-\mathds{1}\{x < q\}\right)(x-q).
    \label{eq:pinball}
\end{equation}
The CRPS has an intuitive definition as the pinball loss integrates over all quantile levels $\alpha \in [0, 1]$, 
\begin{equation}
    \operatorname{CRPS}\left(F^{-1}, x\right)=\int_{0}^{1} 2 \Lambda_{\alpha}\left(F^{-1}(\alpha), x\right) d \alpha\,,
    \label{eq:crps_pinball}
\end{equation}
where $F^{-1}$ represents the quantile function. In practice, we approximate quantiles based on the samples we have. Therefore, equation~\ref{eq:crps_pinball} can be approximated as a summation over N quantile. The precision of our approximation depends on the number of quantiles as well as the number of samples we have.\medskip\\
\paragraph{Sample Estimation:}  Using lemma 2.2 of~\cite{baringhaus2004new} or identity 17 of~\cite{szekely2005new}, we can approximate $\operatorname{CRPS}$ by
\begin{equation}
    \operatorname{CRPS}(F, x)=\mathrm{E}_{F}\left|X-x\right|-\frac{1}{2} \mathrm{E}_{F}\left|X-X^{\prime}\right|\,,
    \label{eq:crps}
\end{equation}
where $X$ and $X^{\prime}$ are independent copies of a random variable
with distribution function F and finite first moment.~\cite{gneiting2007strictly}

\noindent
To investigate the significance of sample size on the accuracy of different approximation methods, we ran a simple experiment. In this experiment, we assumed that the probabilistic model follows a Gaussian distribution with $\mu=0$ and $\sigma=1$. Then, we approximate $CRPS(F,x)$ where $x=0$ with various sample sizes in range $[200,5000]$. Since we know the probabilistic model distribution, we can calculate the value of CRPS analytically, i.e. $CRPS(F,x) \approx 0.2337$.\medskip

\begin{figure}
\begin{minipage}[c][8cm][c]{0.40\textwidth}

\centering
\includegraphics[width=\textwidth]{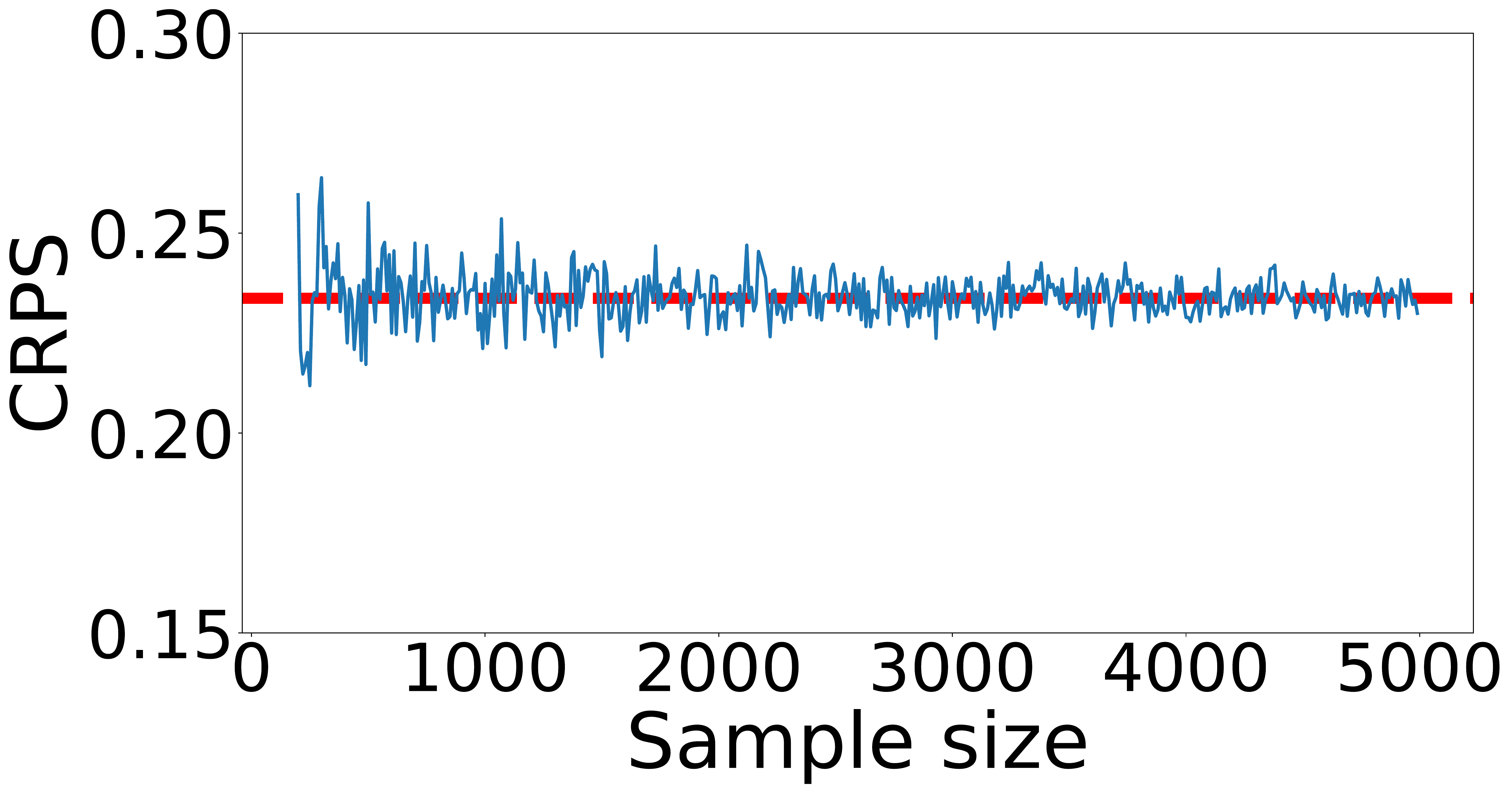}
 \subcaption{Empirical CDF}
 \label{fig:app_emp}

\includegraphics[width=\textwidth]{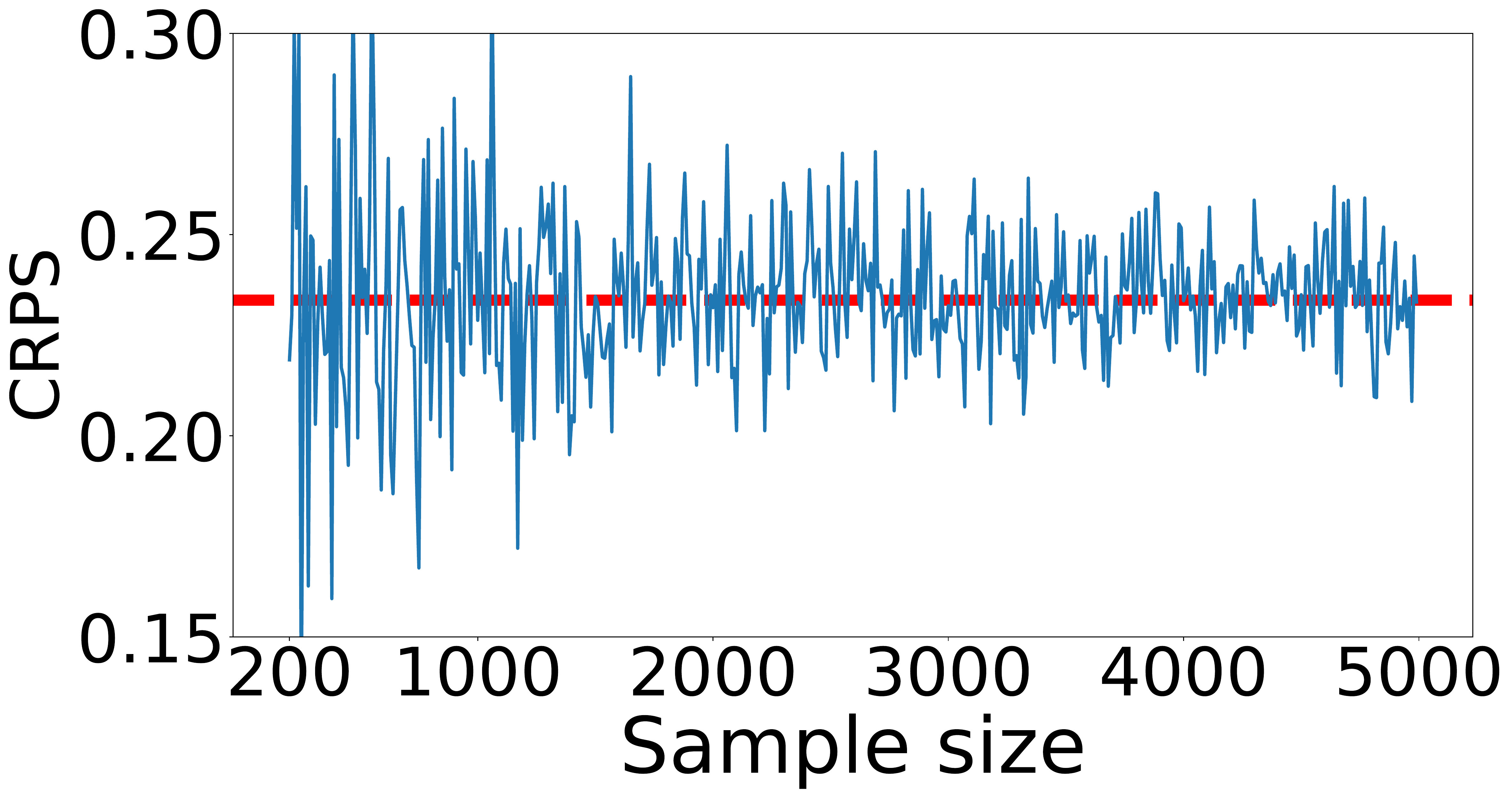}
 \subcaption{Sample estimation}
 \label{fig:app_sam}
\end{minipage}
\begin{minipage}[c][8cm][t]{0.59\textwidth}

\centering
\includegraphics[width=\textwidth]{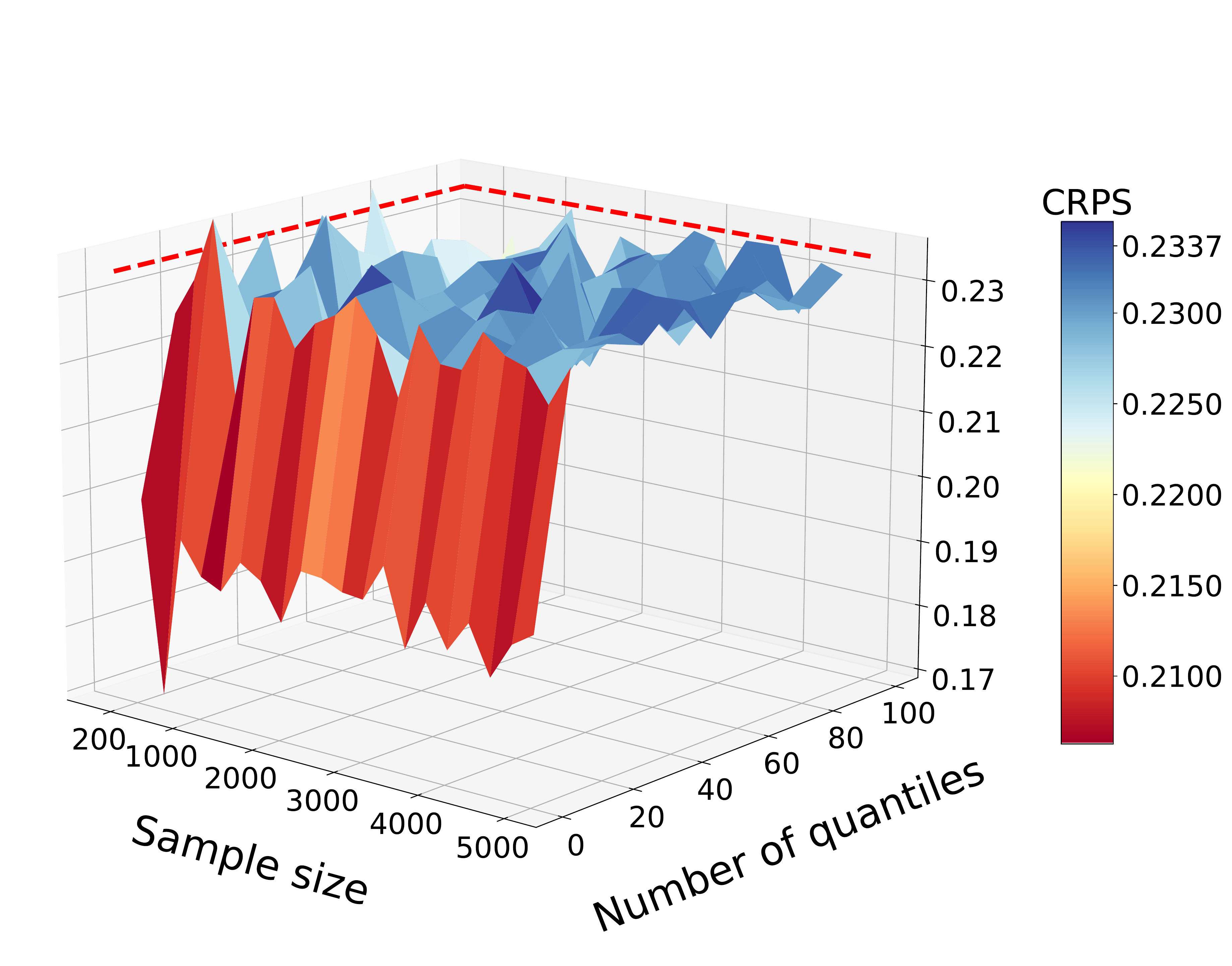}

 \subcaption{Quantile-based}
 \label{fig:app_pin}

\end{minipage}
\vspace{2em}
\caption{The effect of sample size on precision of CRPS approximation using different methods. We can see that all approximation methods can provide us with close estimation, however sample estimation method have more variance in estimation.}
\label{fig:sample}
\end{figure}

\noindent
From figures~\ref{fig:app_emp} and \ref{fig:app_sam}, we can perceive that the empirical CDF method and sample estimation method can converge to the close vicinity of the true value efficiently. However, the empirical CDF method has less variance in comparison to sample estimation. The method based on pinball loss depends on sample size and the number of quantiles. Figure~\ref{fig:app_pin} portrays how these two factors affect the approximation. We can see that with the number of quantiles greater than 20, the pinball loss method can produce a very good approximation using only a few samples (circa 500 samples).

\subsection{Energy Score (ES)}
Energy Score (ES) is a strictly proper scoring rule for multivariate time series. For an m-dimensional observation $x$ in $\mathbb{R}^{m}$ and a predictive cumulative distribution function $F$, the Energy Score (ES)~\cite{gneiting2007strictly} is defined as
\begin{equation}
\mathrm{ES}(F, \mathbf{x})=E_{F}\|\mathbf{X}-\mathbf{x}\|^{\beta} - \frac{1}{2} E_{F}\left\|\mathbf{X}-\mathbf{X}^{\prime}\right\|^{\beta}\,,
\end{equation}
where $\left\|~.~\right\|$ denotes Euclidean distance and $\beta \in (0,2)$. We can see here that CRPS is a special case of ES where $\beta = 1$ and $ m = 1$. While ES is a strictly proper scoring rule for all choices of $\beta$, the standard choice in application is normally $\beta = 1$~\cite{gneiting2007strictly}.\medskip

\noindent
ES provides a method for probabilistic forecast model assessment which works well on multi-variate time series. Unfortunately, ES suffers from the curse of dimensionality~\cite{pinson2013discrimination}and its discrimination power decreases with increasing number of data dimensions. Still, the performance of ES in lower dimensionalities complies with the expected behavior of an honest and careful assessor. Hence, we can use its behavior in lower dimensionalities as the reference for comparison with newly suggested assessment methods.

\subsection{CRPS-Sum}
To address the limitation of ES in multidimensional data, Salinas et al.~\cite{salinas2019high} introduced CRPS-Sum for evaluating a multivariate probabilistic forecasting model. CRPS-Sum is a proper scoring rule, and it is not a strictly proper. CRPS-Sum extends CRPS to multivariate time series with a simple modification. It is defined as
\begin{equation}
\operatorname{CRPS-Sum} = \mathrm{E}_{t}\left[\mathrm{CRPS}\left(F^{-1}_{sum}, \sum_{i} x^{i}_{t})\right)\right]\,,
\label{eq:crps_sum}
\end{equation}
where $F^{-1}_{sum}$ is calculated by summing samples across dimensions and then sorting to get quantiles. Equation~\ref{eq:crps_sum} calculates CRPS based on quantile-based method (equation~\ref{eq:crps_pinball}). In a more general sense, one can calculate the CRPS-Sum by summing both samples and observation across the dimensions. This way, we would acquire a univariate vector of samples and observation. Then, we can apply any aforementioned approximating methods to calculate CRPS-Sum.

\section{Investigating CRPS-Sum properties}
\label{sec:3}
The CRPS-Sum has been wide welcomed by the scientific community and many researches have used it to report the performance of their models~\cite{salinas2019high,rasul2020multi,rasul2021autoregressive,de2020normalizing}. However, the capabilities of CRPS-Sum have not been investigated thoroughly unlike the vast studied dedicated to the properties of ES and CRPS~\cite{gneiting2007strictly,pinson2013discrimination,ziel2019multivariate}. To evaluate discrimination ability of CRPS-Sum, we conducted several experiments on toy dataset and outline the results in this section.

\subsection{CRPS-Sum Sensitivity Study}
In this study, we inspect the sensitiveness of CRPS-Sum concerning the changes in the covariance matrix. This study extends the sensitivity study which were previously conducted by~\cite{pinson2013discrimination,ziel2019multivariate} for various scoring rules including CRPS and ES.
For easier interpretation of the scoring rule response to the changes in the model or data, we define relative changes of the scoring rule $\Delta_{rel}$.\medskip

\noindent
We run our experiment $N$ times, where $CS_{i}$ denotes the obtained CRPS-Sum from the  $i^\text{th}$ experiment. We define
\begin{align}
    \overline{\mathrm{CS}}=\frac{1}{N} \sum_{i=1}^{N} \mathrm{CS}_{i}\,,
\end{align}
as the mean value of CRPS-Sum for the N experiments. Furthermore, let $CS^*$ signify the CRPS-Sum for a model which is identical to the true data distribution. Now, the relative changes~\cite{pinson2013discrimination} in CRPS-Sum is defined as
\begin{equation}
    \Delta_\text{rel}(\mathrm{CS})=\frac{\overline{\mathrm{CS}}-\overline{\mathrm{CS}}^{*}}{\overline{\mathrm{CS}}^{*}}\,.
\end{equation}
\medskip

\noindent
This metric frames the relative changes in CRPS-Sum of a forecasting modeling across our experiments as the differences between the predicted and actual density of the stochastic process. The main idea is to determine the sensitivity of the scores with respect to some biased non-optimal forecast in a relative manner.\medskip\medskip

\noindent
In this study, we have a true data distribution that follows a bivariate normal distribution with $\mu = \begin{pmatrix}
 0\\ 
 0
\end{pmatrix}$ and $\Sigma = \begin{pmatrix}
  ~1~ & ~\rho~\\ 
  ~\rho~ & ~1~
\end{pmatrix}$
where $\rho\in[-1,-0.8,...,0.8,1]$. Furthermore, we specify a forecasting model $f$ which follows the same distribution, however, this time the off-diagonal element of the covariance matrix is $\varrho\in[-1,-0.9,-0.8,...,0.8,0.9,1]$. In our study, we sample $n=2^{14}$ windows of size $w=2^9$ as suggested in~\cite{ziel2019multivariate}.
\begin{figure}
     \centering
     \begin{subfigure}[b]{0.42\textwidth}
         \centering
         \includegraphics[width=\textwidth]{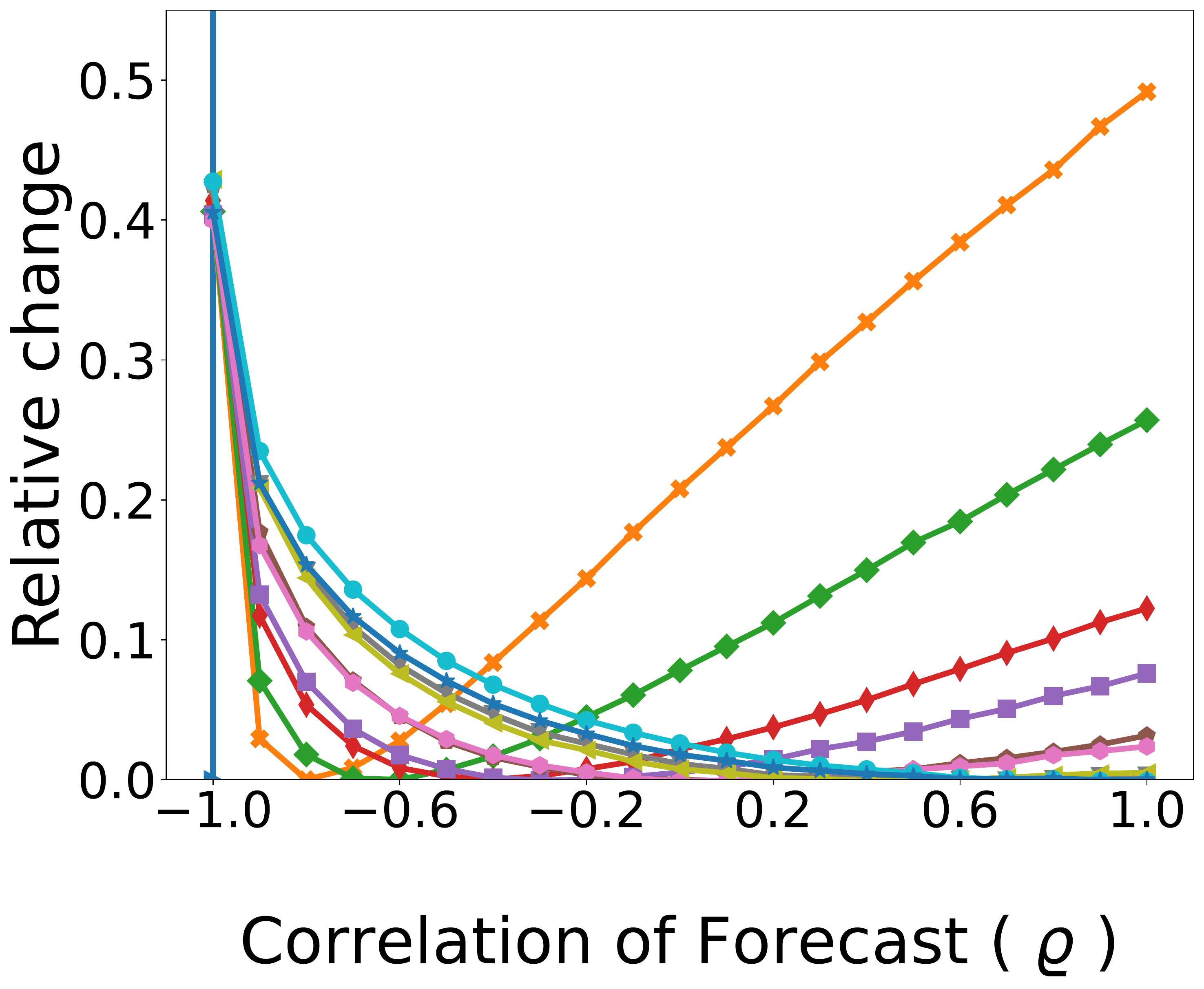}
         \caption{CRPS-Sum}
         \label{fig:rc_cs}
     \end{subfigure}
     \hfill
     \begin{subfigure}[b]{0.57\textwidth}
         \centering
         \includegraphics[width=\textwidth]{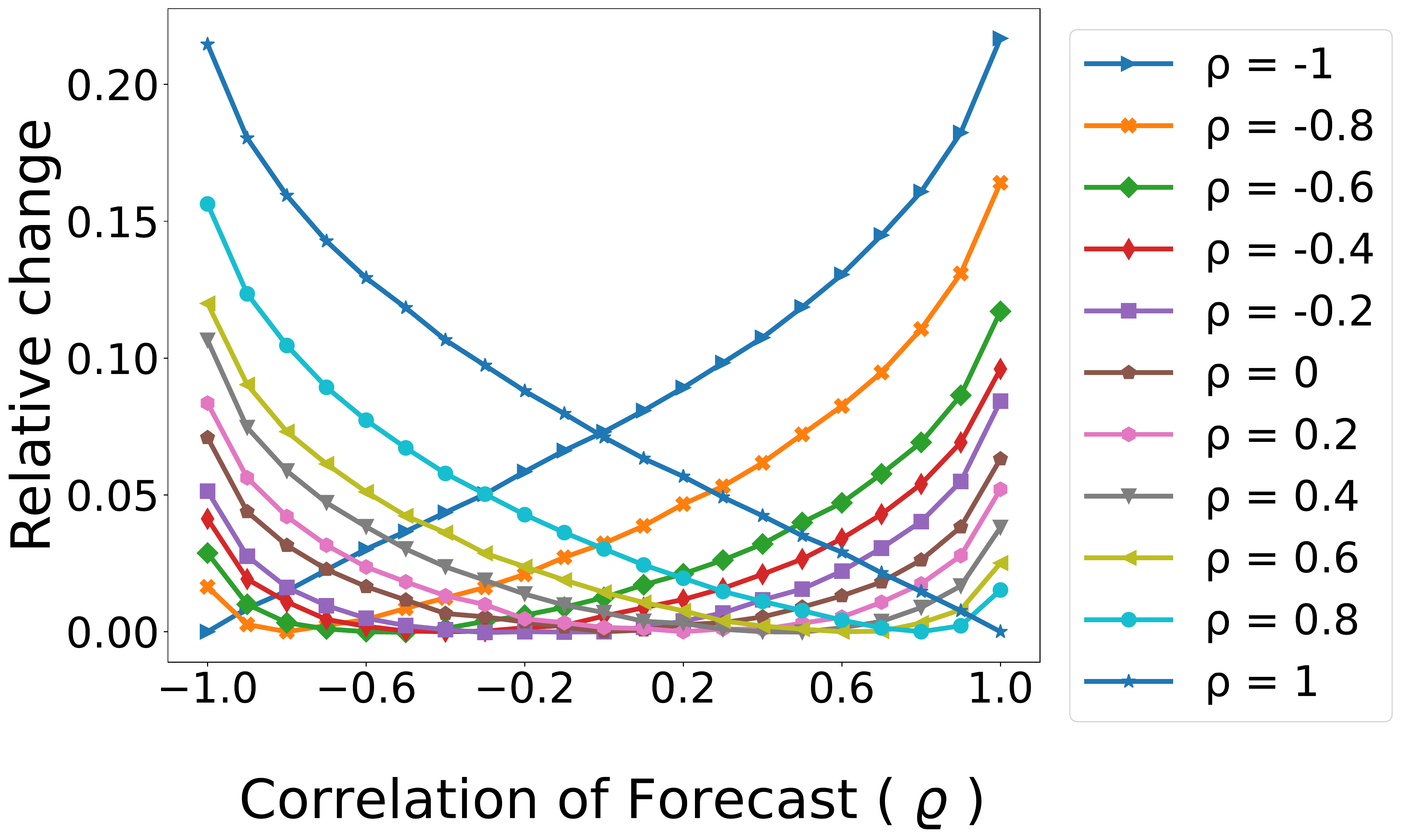}
         \caption{ES}
         \label{fig:rc_es}
     \end{subfigure}
        \caption{The relative change of CRPS-Sum and ES with respect to $\rho$ and $\varrho$. The correlation of forecast($\varrho$) is presented on the x-axis, and the correlation of data($\rho$) is depicted with different lines. Unlike ES, the CRPS-Sum figure is not symmetric, which indicates that it is biased with regard to the $\rho$ value.}
        \label{fig:rel_change}
\end{figure}

\noindent
Figure~\ref{fig:rel_change} illustrates the relative change in CRPS-Sum and ES with respect to changes in correlation $\rho$ of the data generating process as a function of the correlation coefficient $\varrho$ of the family of models we study. We can observe that ES behavior is unbiased with regard to $\rho$ and its figure is symmetric. This is the expected behavior from a scoring rule in this scenario. In contrast, CRPS-Sum response to change in $\rho$ is not symmetric. It is more sensitive to the changes when the covariance $\rho$ of the data is negative, and it is almost indifferent to the changes when the covariance $\rho$ of the data is positive.\medskip

\noindent
Hence, the sensitivity of CRPS-Sum to changes in covariance is dependent on the dependency structure of true data. In real-world scenarios, where we do not have access to the covariance matrix of the data generative process, we cannot reliably interpret CRPS-Sum and compare various models based on CRPS-Sum.

\subsection{The Effect of Summation in CRPS-Sum}
To calculate CRPS-Sum, first we sum the time-series over the dimensions~\cite{salinas2019high}. Although this aggregation let us turn a multivariate time-series into a univariate one, we lose important information concerning the performance of the model on each dimension. Furthermore, the values of dimensions which are negatively correlated, negate each other and consequently those dimension will not be presented in aggregated time series.\medskip

\noindent
For instance, assume we have a multivariate time series $x$ with two dimensions. Our data follow a bivariate Gaussian distribution with 
$\mu = \begin{pmatrix}
 0\\ 
 0
\end{pmatrix}$ and $\Sigma = \begin{pmatrix}
  1 & -1\\ 
  -1 & 1
\end{pmatrix}$.
Hence, the following relation holds between dimensions: 
\begin{equation}
    x^0 = -x^1.
\end{equation}
By summing over dimensions, we have:
\begin{align}
    \sum\limits_i x^i &= 0 \,.
\end{align}
Clearly, after summation we acquire a signal with constant zero and all the information regarding variability of original time series is lost.\medskip

\noindent
To acquire information regarding the performance of the model on each dimension, we can calculate CRPS first. Once the CRPS is validated, we can calculate the CRPS-Sum to check how well the model has learned the relationship between the dimensions, and even at this point, we should not forget the flaws of CRPS-Sum which we witnessed e.g. sensitivity toward data covariance and loss of information during summation. Unfortunately, the importance of CRPS is ignored in most of the recent papers in the probabilistic forecast domain. In these papers, the CRPS is either not reported at all~\cite{de2020normalizing,rasul2021autoregressive}, or the argument about the performance of the model is made solely based on CRPS-Sum~\cite{rasul2020multi,salinas2019high}. Considering the flaws of CRPS-Sum, this trend can put the assessment results of these recent models into jeopardy.

\section{Closer look to CRPS-Sum in Practice}
\label{sec:4}
In previous section, we have discussed the properties of CRPS-Sum and indicate its in  hypothetical scenarios using toy  data settings. In this section, we aim to investigate CRPS-Sum capabilities with a real dataset. To do so, we conduct an experiment by constructing simple models which are based on random noise and investigate its performance using CRPS-Sum. In our experiment, we employed the exchange-rate dataset~\cite{lai2018modeling}. The exchange-rate dataset is a multivariate time series dataset which contains daily exchange-rate of eight countries namely Australia, British, Canada, Switzerland, China, Japan, New Zealand, and Singapore which is collected between 1990 and 2016. This dataset has few dimensions, which let us use ES alongside CRPS and CRPS-Sum. Additionally, it is easier to perform qualitative assessment on lower dimensionalities. We used the dataset in the same setting which is proposed in~\cite{salinas2019high}. We also utilize the GP-Copula method from~\cite{salinas2019high}  as the baseline since they have reported results in CRPS-Sum.\medskip

\noindent
Our first model is a dummy univariate model which follow a Gaussian distribution. The mean of the Gaussian distribution is $\mu = \mu_{last}$ where $\mu_{last}$ is the mean of the last values in the input vector over the dimensions i.e.
\begin{equation}
    \mu_{last} = \frac{1}{D}\sum_{i=1}^{D} x_{T}^{i}.
\end{equation}
We used $\sigma = 10^{-4}$ as the standard deviation of the dummy univariate model in our experiments,  however the results are not dependent on $\sigma$ value (More discussion on $\sigma$ value can be found in Appendix~\ref{appendix:b}). We use this model to generate the forecast for every dimension. \medskip

\noindent
For the second model, we employ a multivariate Gaussian distribution to build a dummy multivariate forecaster. The mean of $ith$ dimension of the multivariate Gaussian distribution is the value of the last time step in the input window, i.e. $\mu_{i} = x_{T}^{i}$. The covariance matrix is zero everywhere except on its main diagonal, which is filled with $10^{-4}$. In other words, we extend the last observation of the input window as the prediction and apply small perturbation from a Gaussian distribution. 

\begin{table}[!h]
\begin{tabular*}{\textwidth}{c @{\extracolsep{\fill}} cccc}
\toprule
         & GP-Copula~\cite{salinas2019high} & Dummy univariate & Dummy multivariate \\ \midrule
CRPS-Sum & 0.0070     & 0.0049           & 0.0048             \\
CRPS     & 0.0092     & 0.4425          & 0.0077           \\
ES       & 0.0043     & 0.2037           & 0.0032             \\ \bottomrule
 & & & \\

\end{tabular*}
\caption{This table illustrates the results from dummy models on the exchange-rate dataset and compares their performance with GP-Copula~(SotA)~\cite{salinas2019high} based on CRPS-Sum, CRPS and ES. It shows that as per CRPS-sum Dummy models have better performance in comparison to GP-Copula. All scoring rules are estimated in similar setting to~\cite{salinas2019high}.}
\label{tab:results}
\end{table}

\noindent
Table~\ref{tab:results} present CRPS-Sum, CRPS and ES of the two dummy models and the result of GP-Copula model from~\cite{salinas2019high} on the exchange-rate dataset. While the CRPS-Sum suggests that dummy univariate model is much better than GP-Copula, the CRPS and ES indicate that the performance of dummy univariate model is worse than GP-Copula. The results reported by CRPS and ES are aligned with our expectations, however, the CRPS-Sum reports a misleading assessment.\medskip

\noindent
On the other hand, the quantitative results for dummy multivariate model are quite surprising. All three assessment methods denote that the dummy multivariate has a superior performance in comparison to GP-Copula. To provide further explanation for this unexpected result, we analyze the performance of these models qualitatively.

\begin{figure}[!h]
    \centering
    \begin{subfigure}{\textwidth}
        \centering
        \includegraphics[width=\textwidth]{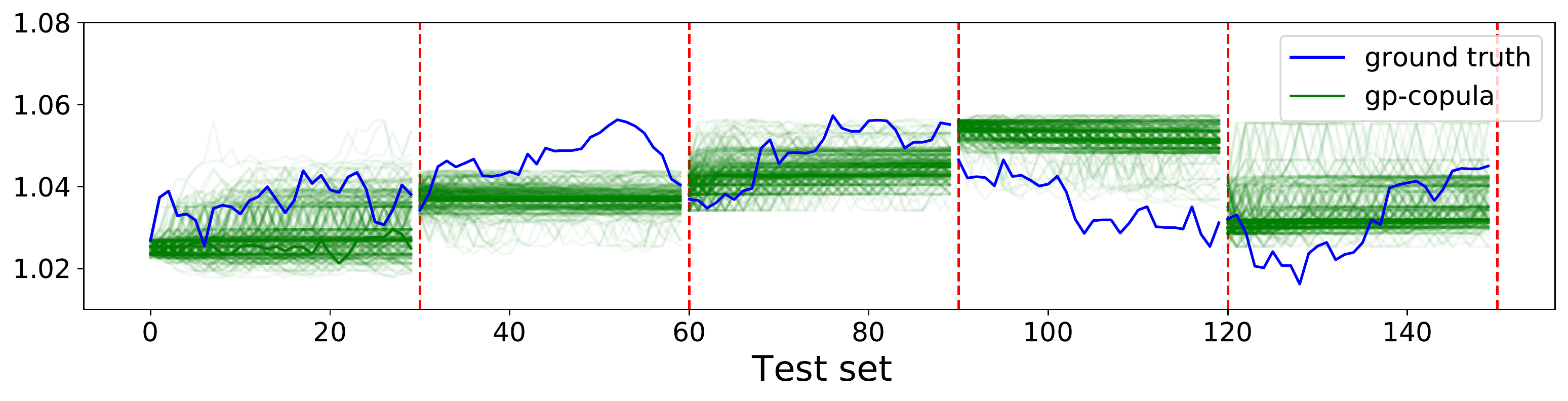}
        \caption{Visualization of GP-Copula~\cite{salinas2019high} samples}
        \label{fig:gp_exchange_sample}
    \end{subfigure}
    
    \hfill
    
    \begin{subfigure}{\textwidth}
        \centering
        \includegraphics[width=\textwidth]{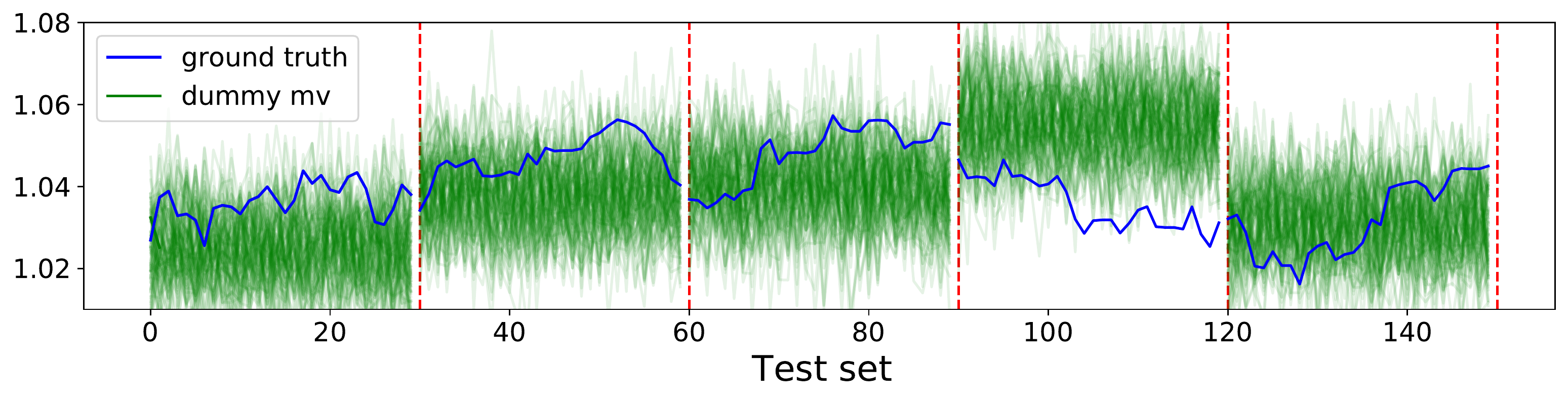}
        \caption{Visualization of forecast samples from dummy multivariate model}
        \label{fig:dummy_exchange_sample}
    \end{subfigure}
    
    \caption{These figures illustrate the 400 forecast samples from GP-Copula and dummy multivariate model for the first dimension of the exchange-rate dataset alongside the expected ground truth. While dummy multivariate model forecasts look like random noise as it is expected, it is hard to spot any meaningful pattern in GP-Copula forecasts in comparison to the expected value.}
    \label{fig:my_label}
\end{figure}

\noindent
Figure~\ref{fig:gp_exchange_sample} depicts the forecasts from GP-Copula for the first dimension of exchange-rate dataset (rest of the dimensions are visualized in Appendix~\ref{appendix:a}) and figure~\ref{fig:dummy_exchange_sample} presents samples from dummy multivariate model. \medskip

\noindent
This experiment shows us that the border between a dummy model and a genuine model can get very blurry, if we rely solely on CRPS-Sum. Furthermore, we learn that CRPS and visualization can help us to acquire a better understanding of a model performance. Still, we do not have a reliable assessment method for investigating dependency structure of a model in higher dimensionalities.

\section{Conclusion}
\label{sec:5}
In this paper, we reviewed various existing methods for the assessment of probabilistic forecast models and discussed their advantages and disadvantages. While CRPS is only applicable on univariate models and ES suffers from curse of dimensionality, CRPS-Sum was introduced to help us with assessing multivariate probabilistic forecast models. Unlike CRPS and ES, the properties of CRPS-Sum are not studied in the past. Our Sensitivity study illustrates that the CRPS-Sum behavior is not symmetric with regard to the covariance of data distribution. CRPS-Sum is more sensitive to changes in covariance of the model when the covariance of the data is negative. It is an undesirable behavior and make the result interpretation difficult.\medskip

\noindent
Furthermore, CRPS-Sum cannot reflect the performance of a model on each dimension due to the loss of information caused by summation during its calculation. We demonstrate this problem with simple examples and experiments on a real world dataset, where a dummy model based on random noise achieved better CRPS-sum than the state-of-the-art model. \medskip

\noindent
To conclude, CRPS-Sum cannot provide an unbiased and accurate assessment for multivariate probabilistic forecasters. Thus, we suggest avoiding CRPS-Sum if possible. For data with low dimensionality, we can use ES. In higher dimensions, the assessment of probabilistic forecast model is still an open problem. In the current state, it is difficult to rely solely on any existing metric and the manual qualitative analysis should be used to evaluate the performance as well. 
\section{Future Works}
\label{sec:6}
Considering the shortcomings of CRPS-Sum, there is an urgent need for an assessment metric for multivariate probabilistic forecast models. A desirable metric would be a strictly proper scoring rule which summarizes the model performance in a single value using a reasonable number of samples. Furthermore, it should be capable of reflecting the precision of the model in learning the probability distribution of each dimension, as well as model accuracy in capturing cross-dimension dependencies.

%
%
%
\printbibliography

\newpage
\begin{subappendices}
\renewcommand{\thesection}{\Alph{section}}%

\section{Visualization of forecasts from GP-Copula on exchange-rate dataset}
\label{appendix:a}

\begin{figure}[H]
    \vspace{-2em}
    \centering
    \begin{subfigure}{0.9\textwidth}
        \centering
        \includegraphics[width=\textwidth]{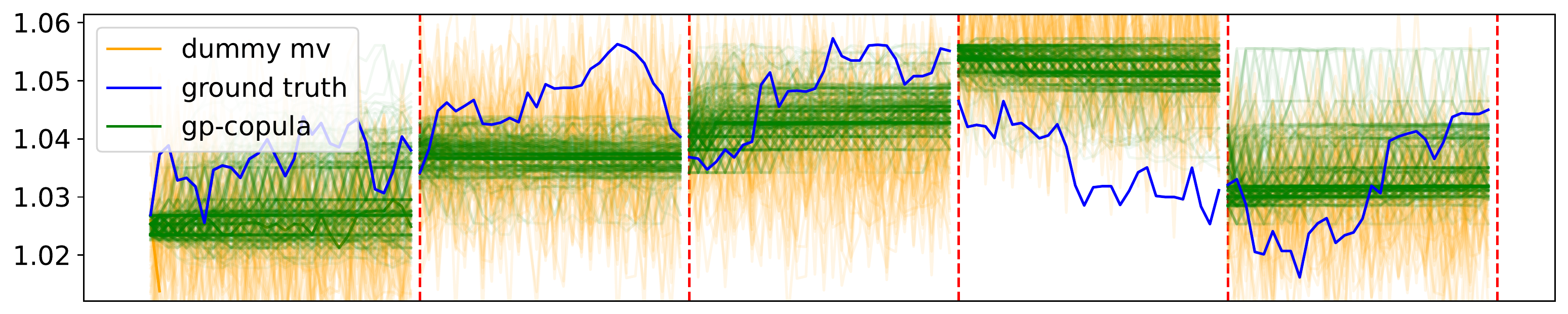}
    \end{subfigure}
    
    \hfill
    
    \begin{subfigure}{0.9\textwidth}
        \centering
        \includegraphics[width=\textwidth]{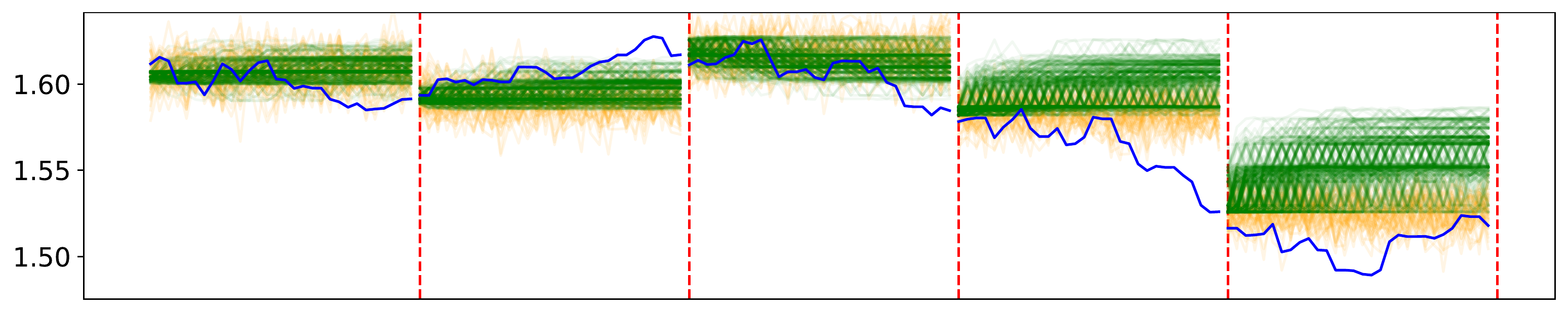}
    \end{subfigure}
    
    \hfill
    
    \begin{subfigure}{0.9\textwidth}
        \centering
        \includegraphics[width=\textwidth]{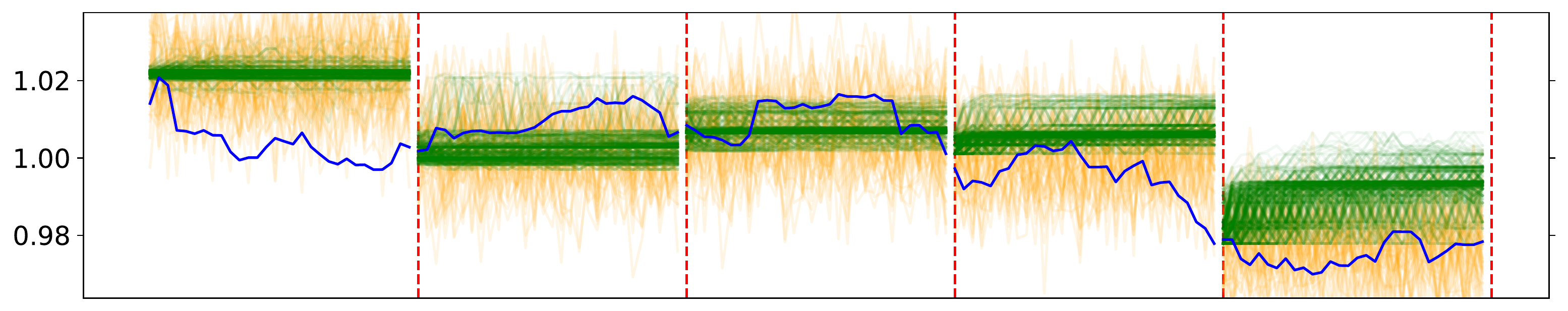}
    \end{subfigure}
    
    \hfill
    
    \begin{subfigure}{0.9\textwidth}
        \centering
        \includegraphics[width=\textwidth]{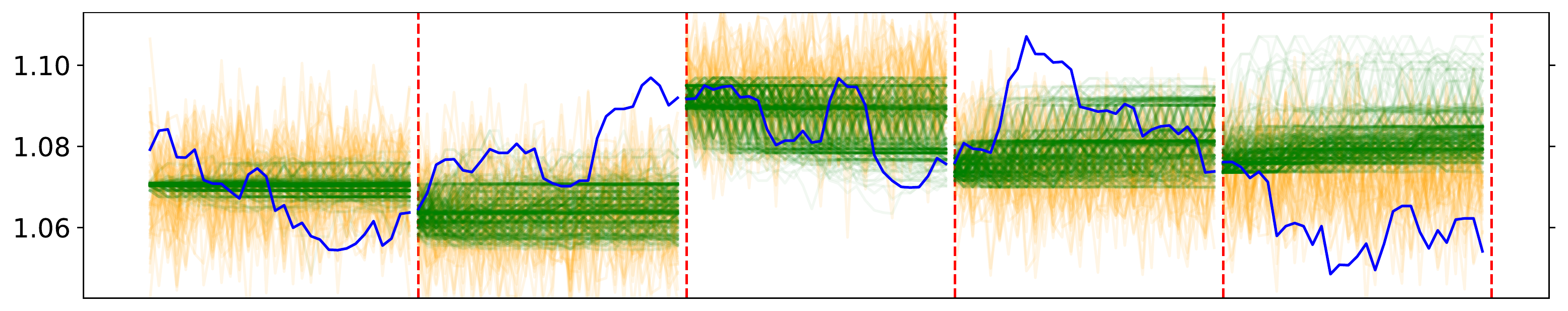}
    \end{subfigure}
    
    \hfill
    
    \begin{subfigure}{0.9\textwidth}
        \centering
        \includegraphics[width=\textwidth]{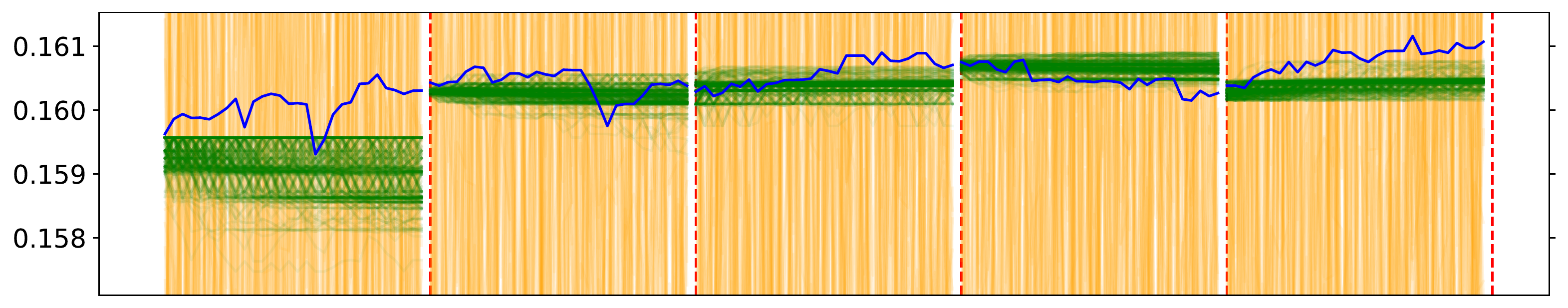}
    \end{subfigure}
    
    \hfill
    
    \begin{subfigure}{0.9\textwidth}
        \centering
        \includegraphics[width=\textwidth]{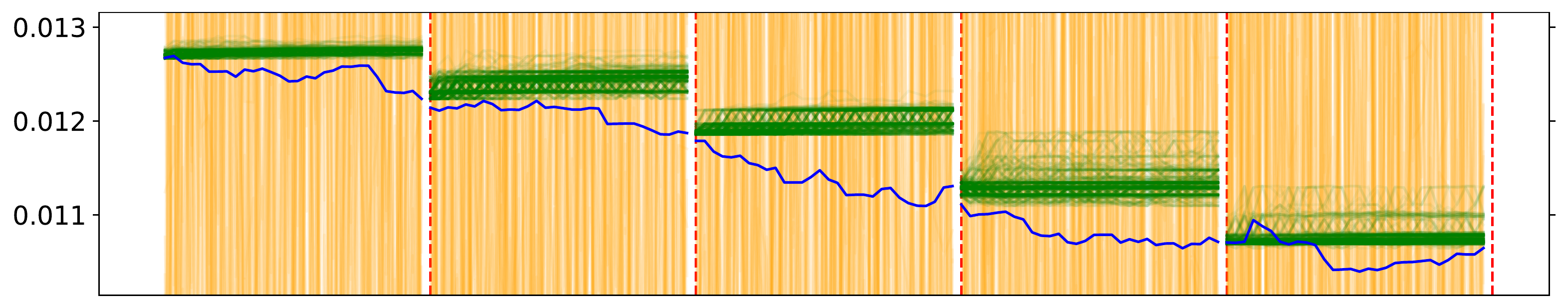}
    \end{subfigure}
    
    \hfill
    
    \begin{subfigure}{0.9\textwidth}
        \centering
        \includegraphics[width=\textwidth]{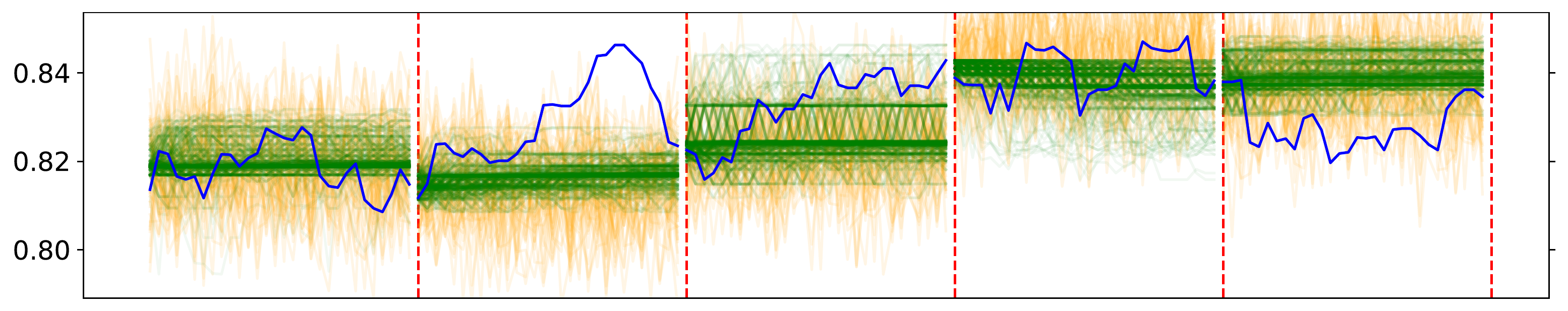}
    \end{subfigure}
    
    \hfill
    
    \begin{subfigure}{0.9\textwidth}
        \centering
        \includegraphics[width=\textwidth]{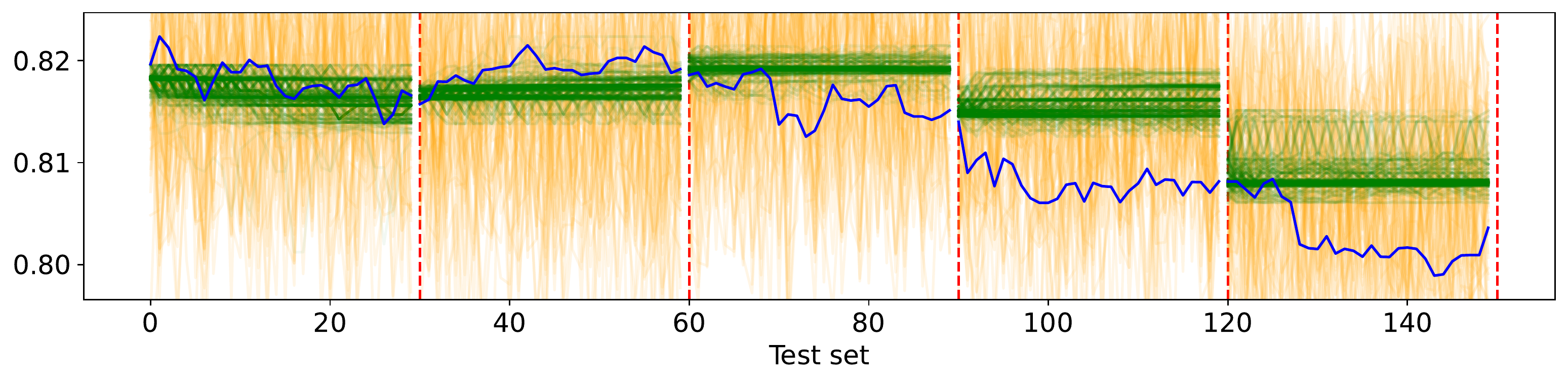}
    \end{subfigure}

    \caption{This figure presents the sample forecasts from GP-Copula~\cite{salinas2019high} for exchange-rate dataset test set. The dataset has 8 dimensions and the test set consists of five batches with 30 time-steps (for more information refer to~\cite{salinas2019high}) . Each sub-figure corresponds to one of the data dimensions, presented in original order from the top to bottom. We used 400 samples for visualization of each forecast batch.}
    \label{fig:gp_exchange}
\end{figure}

\section{The Standard Deviation of Dummy Models}
\label{appendix:b}

\begin{figure}
     \centering
     \begin{subfigure}[b]{0.325\textwidth}
         \centering
         \includegraphics[width=\textwidth]{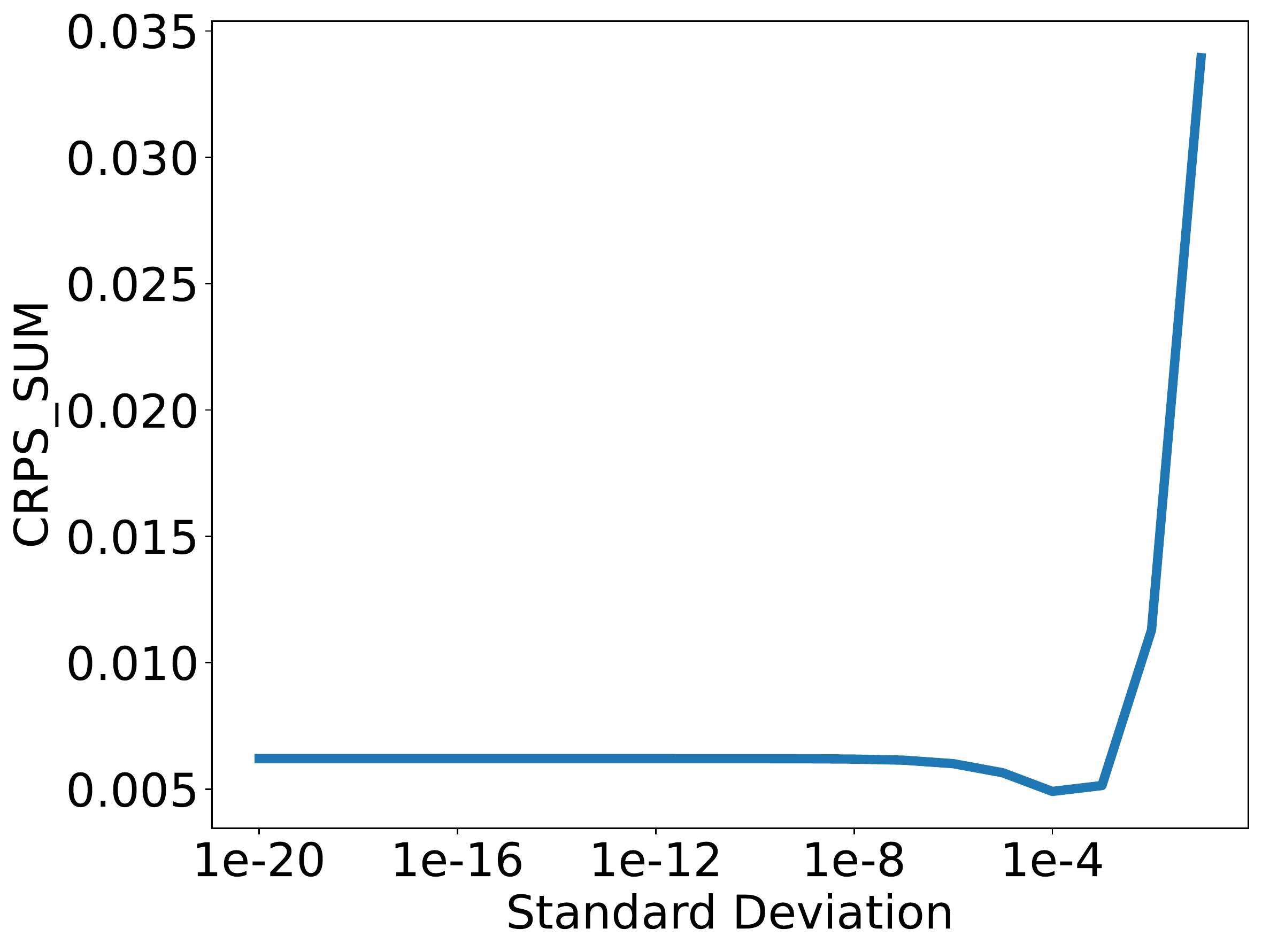}
         \caption{CRPS-Sum}
     \end{subfigure}
     \hfill
     \begin{subfigure}[b]{0.325\textwidth}
         \centering
         \includegraphics[width=\textwidth]{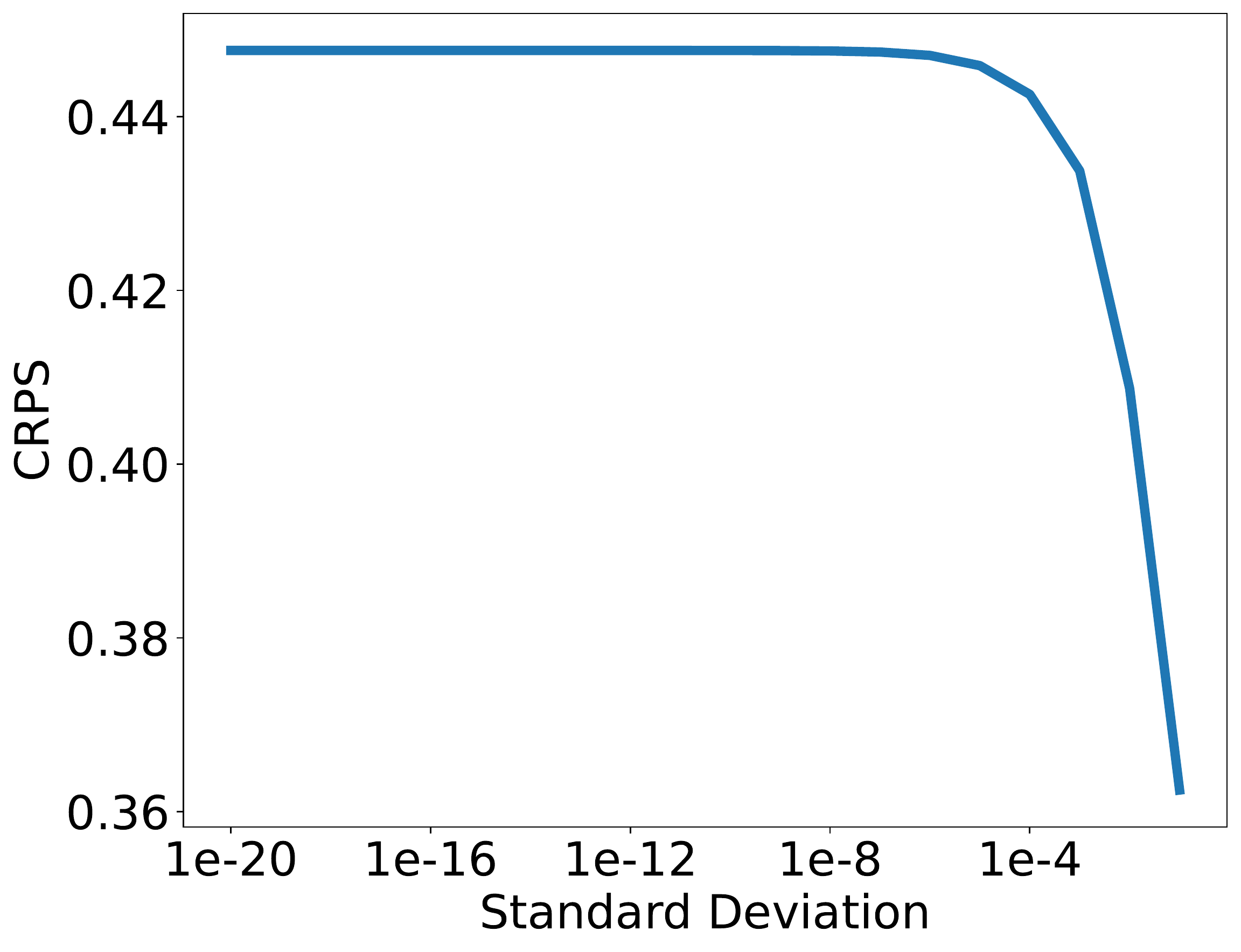}
         \caption{CRPS}
     \end{subfigure}
     \hfill
     \begin{subfigure}[b]{0.325\textwidth}
         \centering
         \includegraphics[width=\textwidth]{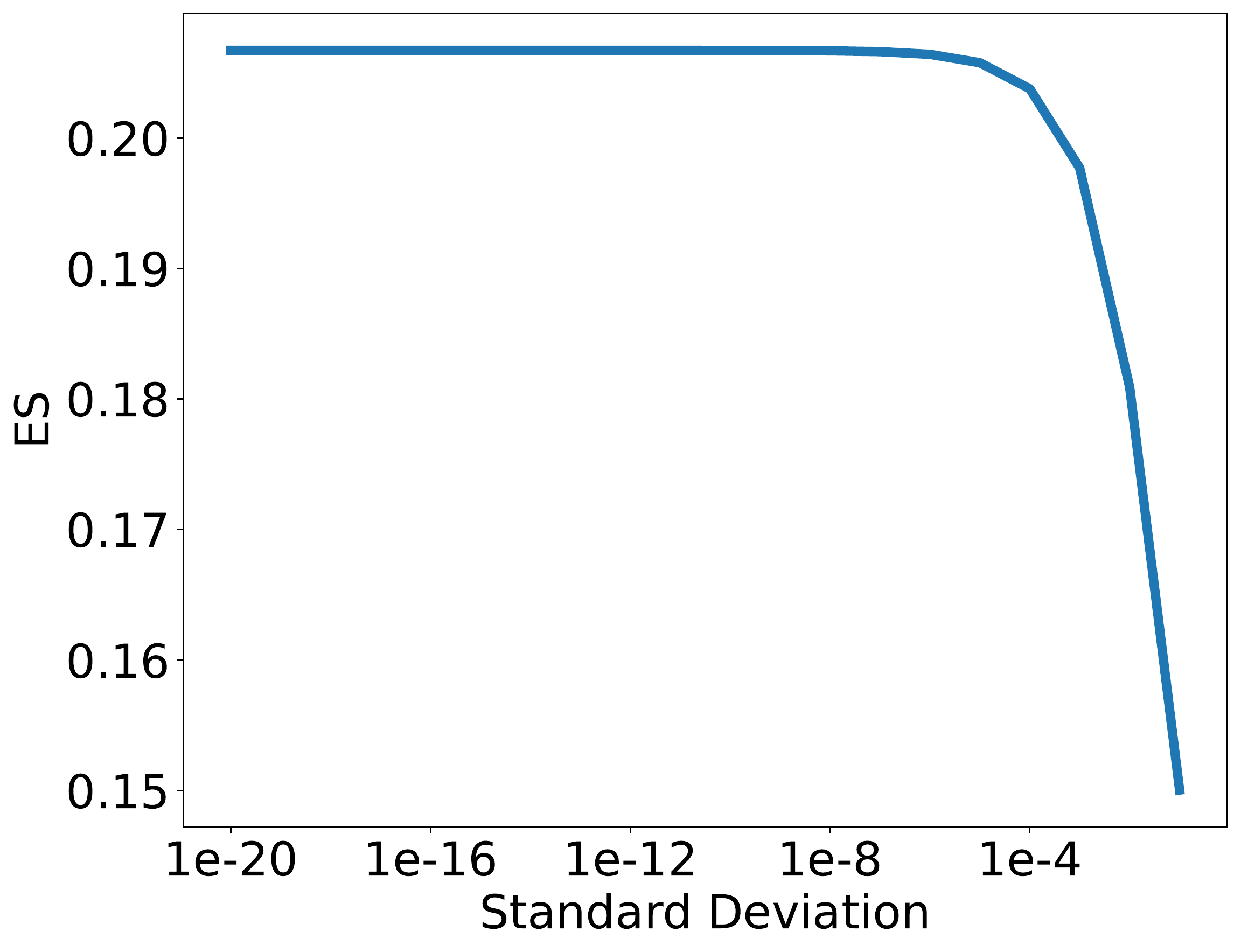}
         \caption{ES}
     \end{subfigure}
        \caption{The assessment of univariate dummy model with $\sigma\in\{1e-1,1e-2,...,1e-20\}$ using CRPS-Sum, CRPS and ES. The plot is depicted on logarithmic scale.}
        \label{fig:std_uni}
\end{figure}

\begin{figure}
     \centering
     \begin{subfigure}[b]{0.325\textwidth}
         \centering
         \includegraphics[width=\textwidth]{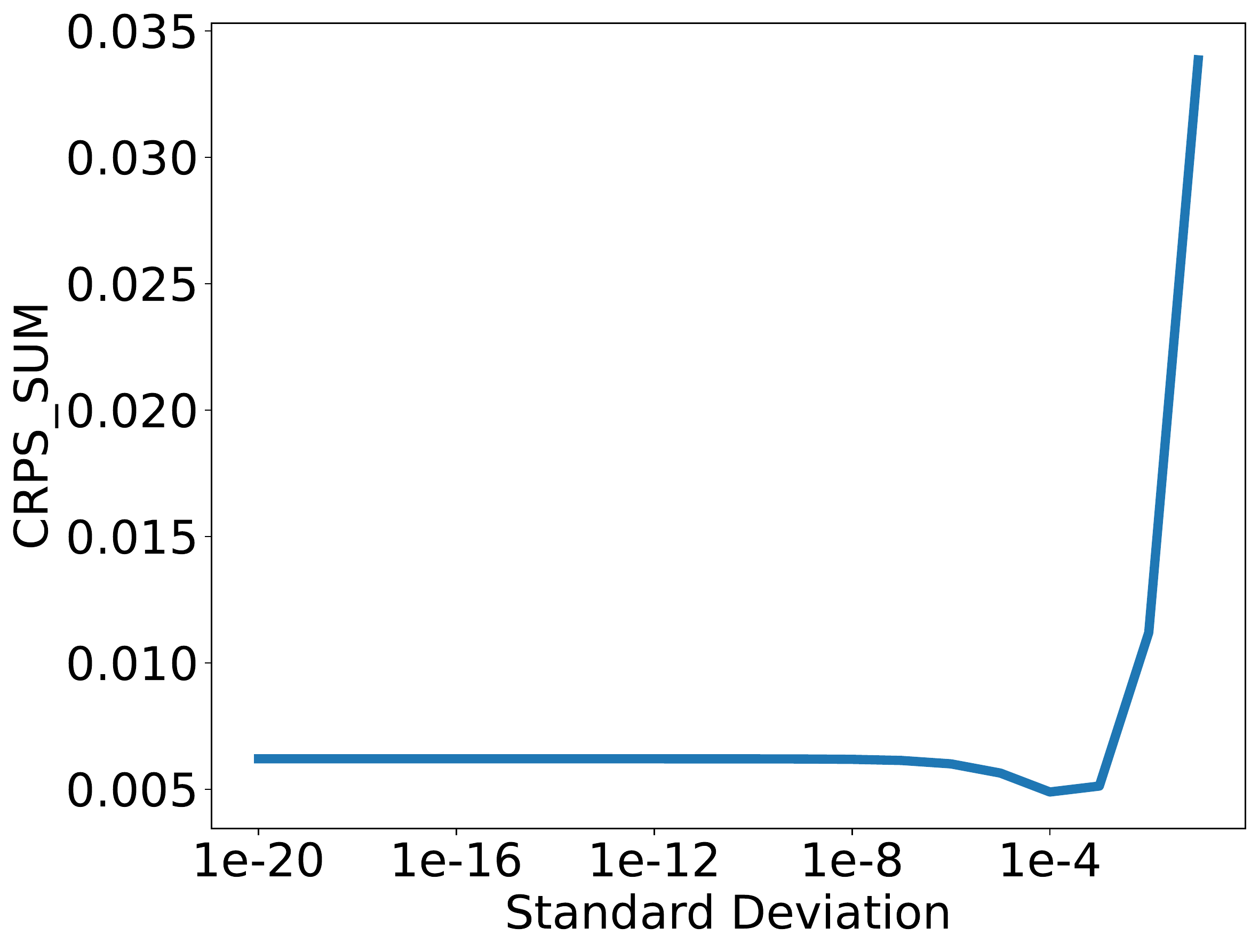}
         \caption{CRPS-Sum}
     \end{subfigure}
     \hfill
     \begin{subfigure}[b]{0.325\textwidth}
         \centering
         \includegraphics[width=\textwidth]{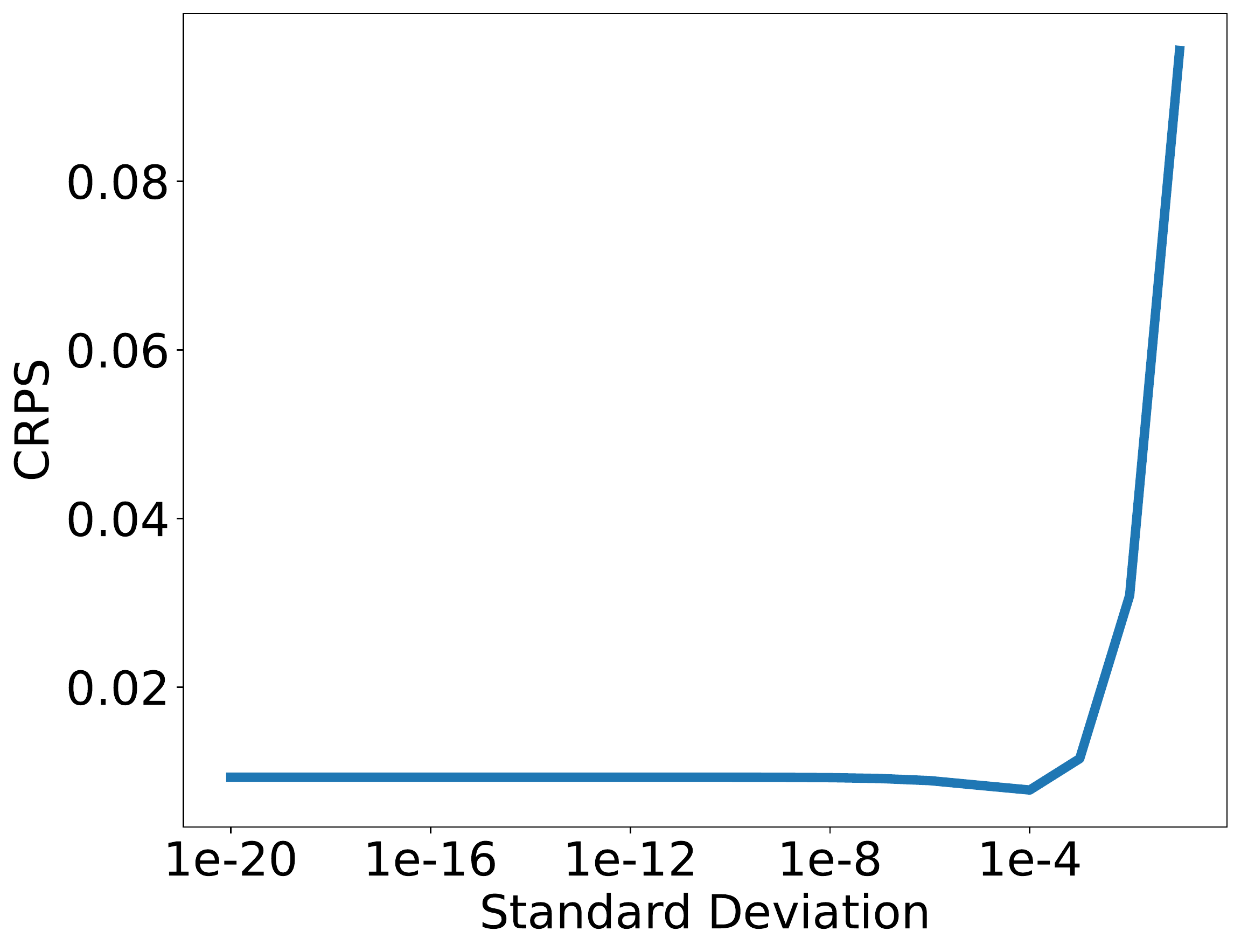}
         \caption{CRPS}
     \end{subfigure}
     \hfill
     \begin{subfigure}[b]{0.325\textwidth}
         \centering
         \includegraphics[width=\textwidth]{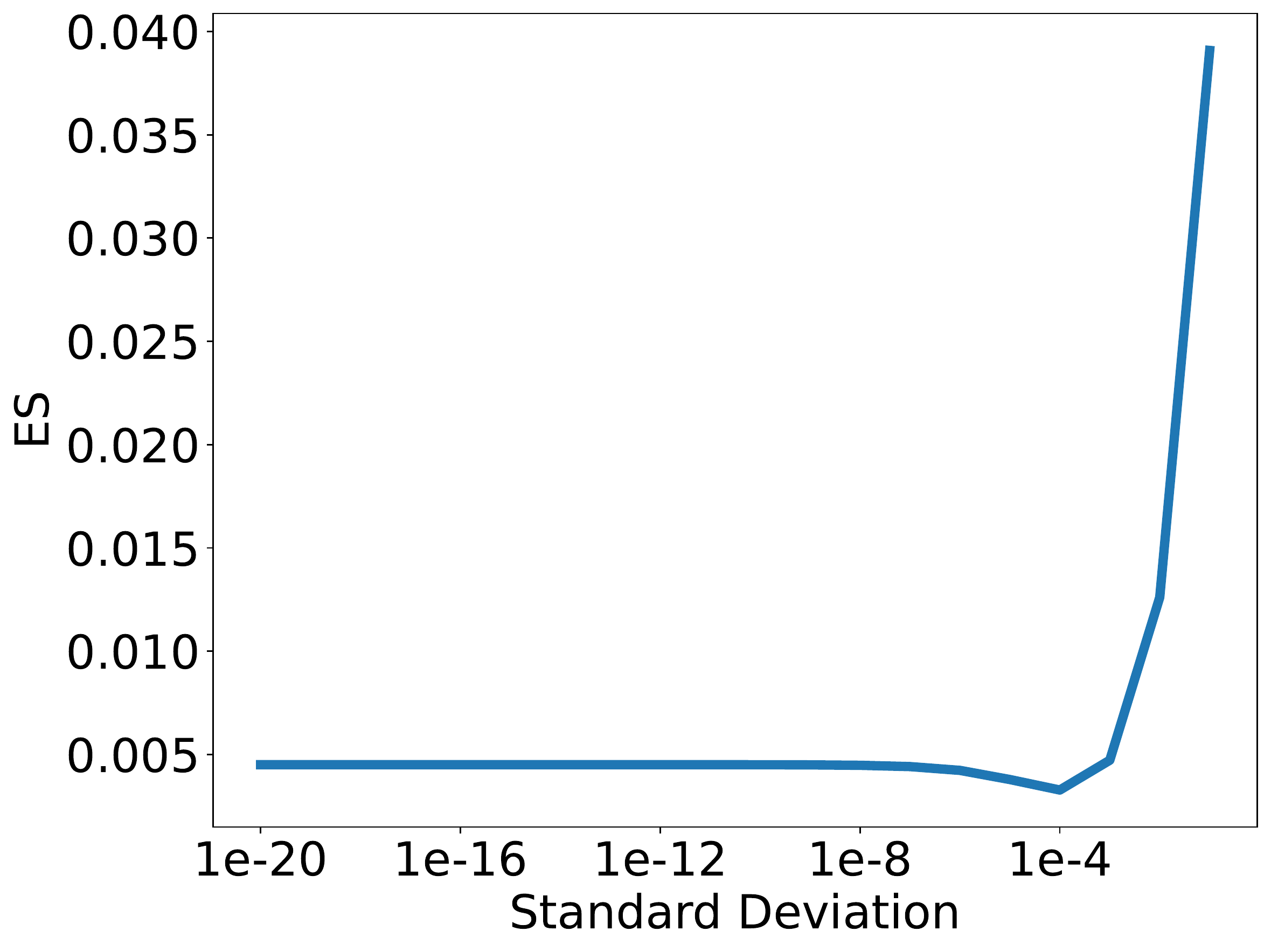}
         \caption{ES}
     \end{subfigure}
        \caption{The assessment of multivariate dummy model with $\sigma\in\{1e-1,1e-2,...,1e-20\}$ using CRPS-Sum, CRPS and ES. The plot is depicted on logarithmic scale.}
        \label{fig:std_mv}
\end{figure}

For our discussions on dummy models performance, we used $\sigma=10^{-4}$ to define Gaussian distribution. Nevertheless, as shown in figure~\ref{fig:std_uni} and figure~\ref{fig:std_mv}, we can acquire consistent result with $\sigma\leq10^{-3}$. Furthermore, we can see that that values of our scoring rules converges when $\sigma\leq10^{-5}$.

\end{subappendices}

\end{document}